# PSDNet and DPDNet: Efficient Channel Expansion, Depthwise-Pointwise-Depthwise Inverted Bottleneck Block

Guoqing Li [a], Meng Zhang [a,*], Qianru Zhang [a], Ziyang Chen [a], Wenzhao Liu [a], Jiaojie Li [a], Xuzhao Shen [a], Jianjun Li [a], Zhenyu Zhu [a], Chau Yuen [b]

[a] *National ASIC Engineering Technology Research Center, School of Electronics Science and Engineering, Southeast University, Nanjing 210096, P. R. China*

[b] *Singapore University of Technology and Design, Singapore*

**Abstract：** In many real-time applications, the deployment of deep neural networks is constrained by high computational cost and efficient lightweight neural networks are widely concerned. In this paper, we propose that depthwise convolution (DWC) is used to expand the number of channels in a bottleneck block, which is more efficient than $1 \times 1$ convolution. The proposed Pointwise-Standard-Depthwise network (PSDNet) based on channel expansion with DWC has fewer number of parameters, less computational cost and higher accuracy than the corresponding ResNet on CIFAR datasets. To design more efficient lightweight convolutional neural network, Depthwise-Pointwise-Depthwise inverted bottleneck block (DPD block) is proposed and DPDNet is designed by stacking DPD block. Meanwhile, the number of parameters of DPDNet is only about 60% of that of MobileNetV2 for networks with the same number of layers, but it can achieve approximate accuracy. Additionally, two hyperparameters of DPDNet can make the trade-off between accuracy and computational cost, which makes DPDNet suitable for diverse tasks. Furthermore, we find that networks with more DWC layers outperform the networks with more $1 \times 1$ convolution layers, which indicates that extracting spatial information is more important than combining channel information.

1. Introduction

Convolution neural networks (CNNs) have been widely used in visual task [1-3] fields since AlexNet [4] won the ILSVARC-2012 [5] championship. Considering better performance to achieve in more complex tasks, the primary trend is building deeper and wider CNNs like VGG [6], GoogLeNet [7], ResNet [8], DenseNet [9], WRNs [10], ResNeXt [11]. Subsequently, some approaches are proposed for higher accuracy, such as Elastic Rectified Linear Unit (EReLU) [12], multiple convolutional layers fusion [13], boosted convolutional neural network [14] and so on. However, limited computing resources make complex models difficult to be applied in the scenes of robotics, AR, smartphone, .etc. Thereby, reducing computation complexity is of great significance. In traditional machine learning fields, Xu et al. propose an efficient method of crowd aggregation computation in public areas, which can effectively reduce the number of parameters. An efficient shadow recognition approach is proposed, in which a Gaussian model based method is used to reduce parameters. In CNNs, the efficient and lightweight CNN architectures have attracted more attention. In this paper, we main focus on the efficient CNN architecture. A variety of approaches [15] have been proposed for efficient CNN architectures, which can be divided into two main kinds of aspects [16]: compressing existing architecture [17] with pre-trained models and designing new efficient architectures [18, 19] that will be trained from scratch.

The bottleneck block can reduce the parameter redundancy, which is widely adopted in various CNNs [8, 10, 11, 20, 21]. In a bottleneck block, the number of channels is squeezed first and then expanded by pointwise convolution (PWC) which is also named 1×1 convolution. It is well known that the number of channels in CNNs is very large, which results in a huge number of parameters and together with the considerable computational cost of the PWC, especially in the last few convolutional layers. Furthermore, MobilenetV2, an efficient CNN, introduced the inverted bottleneck block and adopted depthwise separable convolution [22]. In an inverted bottleneck block, the number of channels is expanded first and then squeezed. The inverted bottleneck block with depthwise separable convolution is more efficient than the original bottleneck block.

However, previous researches ignored the channel expansion capability of DWC. In this paper, we propose using DWC instead of PWC to expand the number of channels in a bottleneck block, which can reduce the number of parameters and computation complexity. Moreover, a novel efficient CNN architecture PSDNet is proposed, which is more efficient than ResNet. Furthermore, we propose a more efficient DPDNet, which mainly uses DWC, PWC and inverted bottleneck block to reduce the redundancy. DPDNet has two hyperparameters, which can trade off the complexity and accuracy of the model to be applied in different scenarios.

Our main contributions are as follows:

(1) We propose DWC layer as the channel expansion layer in a bottleneck block, which is more efficient than PWC.

(2) We introduce an efficient CNN architecture PSDNet, which has fewer parameters, less computational cost and higher accuracy than ResNet.

(3) We present a novel efficient CNN architecture DPDNet, which is more efficient than MobilenetV2 considering the number of parameters and computational complexity.

(4) We show that CNNs with more DWC layers have better performance than ones with more PWC layers, which indicates extracting spatial features is more effective than combining the information on different channels.

This paper is organized as follows. Section 2 reviews some related work about efficient CNNs and the proposed DPDNet is detailed in Section 3. We present the experimental details and results to evaluate the proposed CNN architectures in Section 4 and finally the conclusion is presented in Section 5.

2. Related Work

In recent years, researchers have shown great interest in efficient lightweight networks. In this section, we are going to briefly review previous tasks which inspire the design of our network. We consider five related tasks which are pruning, quantization, group convolution, small kernels and depthwise separable convolution.

**Pruning.** Network pruning originates as a method to reduce the size and over-fitting of a neural network. The pruned CNN is sparser, which can reduce the overhead of computation and memory resources. Many pruning methods involved in connections [17, 23], filters [24], channels [25] and so on. The pruned CNN is efficient and highly accurate by fine-tuning.

**Quantization.** A method of compressing the CNNs is to use low-bit fixed-point weight, which can keep a competitive performance. For example, Han et al. achieved further storage reduction by means of 8-bit quantization without any loss [17]. Based on binarized-neural-network [26, 27], more efficient XNOR-Net [28] was proposed and accomplished 58× speedup. Ternary CNNs [29, 30] can achieve a balance between accuracy and computational complexity. These binary and ternary CNNs drastically reduce memory consumption and can use bit-wise operations to replace multiplication and accumulation operations, which leads to an increase in power efficiency.

**Group convolution.** Group convolution is first applied to AlexNet [4] to solve the problem of insufficient memory. Recently, some CNNs [11, 19, 31] use group convolution to improve its performance and reduce computational complexity. These CNNs have more channels under the same parameter scale, which leads to better performance and higher efficiency.

**Small kernels.** We know that smaller convolution kernels have fewer number of parameters, which were used a few years ago. Initially, all filters keep the size of 3×3 in VGGNet [6]. Then, 3×3 filters are widely adopted in CNNs [9, 18, 20, 32-34]. Multiple 3×3 convolution kernels with the fewer number of parameters can replace a larger convolution kernel with larger number of parameters. For instance, two 3×3 kernels can replace a 5×5 kernel. Moreover, the strategy of using $1 \times n$ and $n \times 1$ convolution kernels instead of $n \times n$ convolution kernel is also widely used in Flattened networks [35], Inception models [32, 36, 37] and VeckerNets [38]. 1×1 convolution kernel has fewer parameters, which is often used to increase or decrease the number of channels in CNNs. Currently, the small kernel is a very popular method to reduce the number of parameters and various CNNs with small convolution kernels have competitive performances.

**Depthwise separable convolution.** A standard convolution can be decomposed into a DWC and a PWC. DWC can extract the spatial information of one feature map while PWC can assemble the characteristics of all channels. Xception [34] utilizes the depthwise separable convolution and gains high accuracy on ImageNet [5] dataset. MobileNet [18] gains state-of-art results among lightweight CNNs by using depthwise separable convolution. Our work mainly uses depthwise convolution to improve parameter efficiency much further.

## 3. Our Network

In this section, we first compare the difference between DWC and PWC in terms of the number of parameters, computational cost, and their features. Then our network architectures are introduced in detail.

### 3.1 Depthwise and Pointwise Convolution

Standard convolution can be achieved by combining DWC and PWC. Compared to standard filter, depthwise separable convolution filter has fewer parameters. Furthermore, DWC and PWC can also be used separately. We know that 1×1 convolution has fewer parameters than 3×3 standard convolution when increasing or decreasing the number of channels. Therefore, PWC is widely used to increase or decrease the number of channels in CNNs [8, 20-22, 36]. PWC operation is shown in Figure 1 (a). ResNet [20] uses 1×1 convolution to increase and decrease the number of channels so that the number of parameters does not increase explosively. When the DWC is adopted, the channel-multiplier ($m$) is usually set as 1 in most CNNs [18, 22, 34], which means that the number of input channels is the same as the number of output channels. Figure 1 (b) show the DWC ($m = 1$) operation. If $m$ is set as an integer greater than 1, then DWC can also implement channel expansion. DWC operation with channel expansion is shown in Figure 1 (c). In this paper, we propose to use DWC to expand the number of channels.

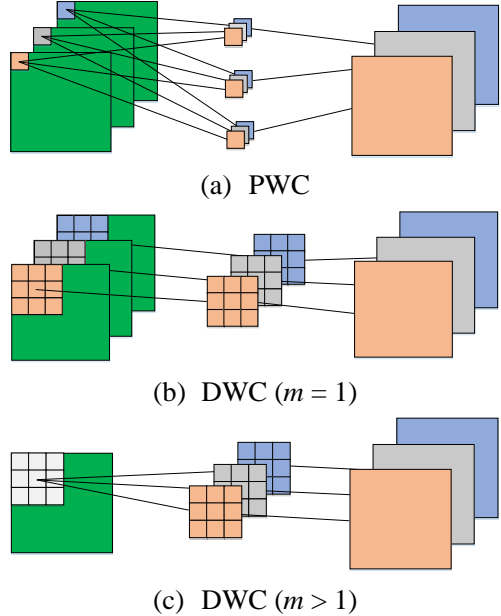

(a) PWC

(b) DWC ($m = 1$)

(c) DWC ($m > 1$)

Figure 1. The illustration of PWC, DWC ($m = 1$), DWC ($m > 1$).

We assume a convolution operation to expand feature map channels. A convolution layer takes a feature map F ($W \times H \times C$) as input and produces a feature map G ($W \times H \times mC$, ($m$ is an integer)). We assume the width and height of output feature map are the same as that of the input feature map, where $W$ and $H$ mean the spatial width and height of the feature map, while $C$ and $mC$ represent the number of input channels and output channels respectively. We assume that the number of output channels is greater than the number of input channels, so $m$ is greater than 1.

To complete the convolution defined above, the number of parameters that PWC filter need is $C \cdot mC$. The computational cost of PWC is $W \cdot H \cdot C \cdot mC$. The size of pointwise convolution kernel is 1, so the PWC has fewer number of parameters and less computational cost than standard convolution (3×3). The number of parameters and computational cost depend on the number of input and output channels and the feature map size. PWC combines the features of different channels and produces new features, but can't filter spatial features of input feature maps.

Comparatively, when DWC completes the convolution operation showed above, the number of parameters is $k \cdot k \cdot mC$, and DWC has a computational cost of $W \cdot H \cdot k \cdot k \cdot mC$. Obviously, DWC is more efficient than standard convolution, because it does not need to combine the information of all channels. The number of parameters depends on the kernel size $k$ and the number of output channels $mC$. The feature map size $W \times H$, kernel size $k$ and the number of output channels $mC$ decide the computational cost. Generally, the DWC applies only one filter ($m = 1$) on each input channel. So the number of input channels is the same as that of output channels. The depthwise convolution can apply multiple filters ($m > 1$) to each input channel, which can extract multiple spatial features on one channel and produce more output channels.

$$\frac{C \cdot mC}{k \cdot k \cdot mC} = \frac{C}{k^2} \tag{1}$$

$$\frac{W \cdot H \cdot C \cdot mC}{W \cdot H \cdot k \cdot k \cdot mC} = \frac{C}{k^2} \tag{2}$$

Eq. (1) and Eq. (2) are the ratios between PWC and DWC in terms of the number of parameters and computational cost, respectively. The ratios of the number of parameters and computational cost are equal. Generally, the kernel size $k$ is 3, so $k^2$ is 9. In most cases, $C$ usually is an integer much greater than 9 in CNNs. Therefore, the ratio $C/k^2$ is much greater than 1. We can find that the DWC has fewer number of parameters and less computational cost than PWC when increasing the number of channels. For this reason, this paper proposes to use DWC to increase the number of channels in CNNs.

### 3.2 Network Architecture

#### 3.2.1 PSDNet

ResNet with bottleneck block has high performance and is efficient in parameter, because of utilizing a large number of 1×1 convolutions to reduce the number of parameters. As shown in Figure 2, in a bottleneck block of ResNet, there are two 1×1 convolution layers and one 3×3 convolution layer. The first 1×1 convolution layer decreases the number of channels and the other one increases the number of channels. The number of parameters of two 1×1 convolution layers are almost equal to the number of parameters of 3×3 convolution layer. The massive parameters are used to combine information on different channels. We propose to replace the $1 \times 1$ convolution for channel expansion with DWC (3×3) in a ResNet block. The new block shown in Figure 2 (b) is called pointwise-standard-depthwise (PSD) bottleneck block. In this paper, ResNet50 is chosen as the base model. The original ResNet50 [20] is designed for ImageNet [5] dataset and the number of its channels is very large. Furthermore, ResNet50 has 5 down sampling layers. The CIFAR datasets are simpler than ImageNet dataset. Therefore, the original ResNet50 is not suitable for CIFAR. We compress the original ResNet50 by reducing the number of channels and the number of down sampling layers. The architecture of compressed ResNet50 is shown in Table1. Then the bottleneck block of compressed ResNet50 is replaced by PSD block, which is called PSDNet50 as shown in Table 1. Down sampling is handled with stride convolution in the 3×3 standard convolutions. From Eq (5) and Eq (6), we can know that PSDNet has fewer number of parameters and less computational cost than ResNet50.

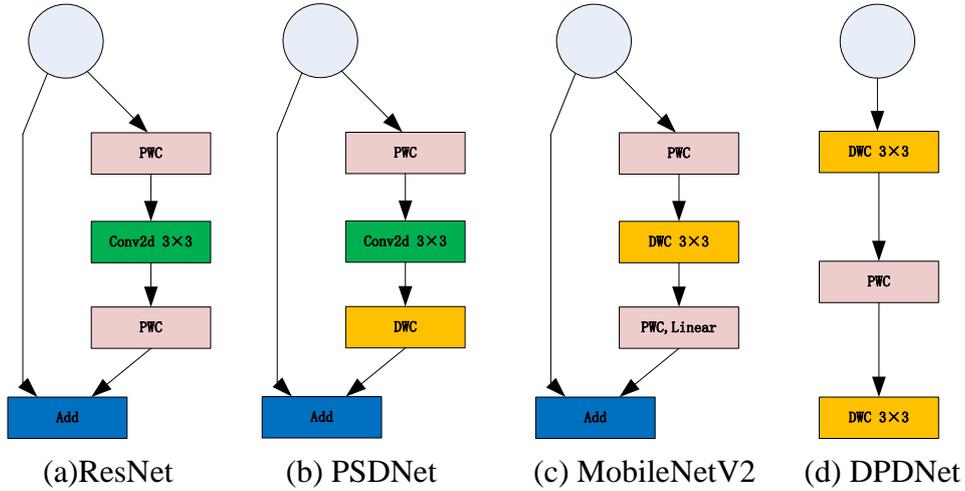

|(a)ResNet | (b) PSDNet | (c) MobileNetV2 | (d) DPDNet|

Figure 2. Comparison of bottleneck blocks for different architectures. The DPDNet block has two DWC layers and one PWC layer, and the first DWC layer implements channel expansion.

Table 1. ResNet50 and PSDNet50 body architectures for CIFAR.

| Output size | ResNet50 | PSDNet50 |
|:---:|:---:|:---:|
| $32 \times 32$ | \multicolumn{2}{c}{$3 \times 3, 16$} |
| $32 \times 32$ | $\begin{bmatrix} 1 \times 1, 16 \\ 3 \times 3, 16 \\ 1 \times 1, 64 \end{bmatrix} \times 5$ | $\begin{bmatrix} 1 \times 1, 16 \\ 3 \times 3, 16 \\ \text{DWC}, 64 \end{bmatrix} \times 5$ |
| $16 \times 16$ | $\begin{bmatrix} 1 \times 1, 32 \\ 3 \times 3, 32 \\ 1 \times 1, 128 \end{bmatrix} \times 6$ | $\begin{bmatrix} 1 \times 1, 32 \\ 3 \times 3, 32 \\ \text{DWC}, 128 \end{bmatrix} \times 6$ |
| $8 \times 8$ | $\begin{bmatrix} 1 \times 1, 64 \\ 3 \times 3, 64 \\ 1 \times 1, 256 \end{bmatrix} \times 5$ | $\begin{bmatrix} 1 \times 1, 64 \\ 3 \times 3, 64 \\ \text{DWC}, 256 \end{bmatrix} \times 5$ |
| $1 \times 1$ | \multicolumn{2}{c}{Avg pool} |
| - | \multicolumn{2}{c}{FC} |
| - | \multicolumn{2}{c}{Softmax} |

### 3.2.2 DPDNet

MobilenetV2 [22] is a very efficient CNN architecture, which takes advantage of depthwise separable convolution. Furthermore, the inverted residuals bottleneck block is proposed in MobilenetV2, which is shown in Figure 2 (c). In an inverted bottleneck, the number of channels is first expanded and then squeezed. Inspired by MobilenetV2, we propose a more efficient inverted bottleneck block. The details of one inverted bottleneck block in our network are shown in Table 2 and Figure 2 (d). In one block, there are three convolution layers and the kernel size of all DWC is 3×3. The first layer is DWC ($m > 1$) layer with stride $s$, which is used to increase the number of channels and complete down sampling. PWC layer is followed, which can combine the information of different channels and decrease number of channels. Finally, the DWC layer ($m = 1$) filters each channel. Our proposed bottleneck block is called depthwise-pointwise-depthwise (DPD) block, which has fewer parameters and less computational cost.

Table 2. The DPD block transforming from $k$ to $k'$ channels, with stride $s$, and channel-multiplier $m$.

| Input | Operator | Output |
|:---:|:---:|:---:|
| $h \times w \times mk$ | DWC ($m > 1$) | $\frac{h}{s} \times \frac{w}{s} \times mk$ |
| $\frac{h}{s} \times \frac{w}{s} \times mk$ | PWC | $\frac{h}{s} \times \frac{w}{s} \times k'$ |
| $\frac{h}{s} \times \frac{w}{s} \times k'$ | DWC ($m = 1$) | $\frac{h}{s} \times \frac{w}{s} \times k'$ |

Table 3. DPDNet Body Architecture for CIFAR 10.

| Output | Operator | $s$ |
|:---:|:---:|:---:|
| 32×32×32 | Conv2d | 1 |
| 32×32×16 | DPD block | 1 |
| 32×32×24 | DPD block | 1 |
| 16×16×32 | DPD block | 2 |
| 16×16×64 | DPD block | 1 |
| 8×8×96 | DPD block | 2 |
| 8×8×160 | DPD block | 1 |
| 1×1×160 | Avg pool $8 \times 8$ | - |
| 1×1×10 | FC | - |
| 1×1×10 | Softmax | - |

By stacking DPD blocks, we construct a novel efficient network called DPDNet for CIFAR dataset, which is shown in Table 3. The first layer is a standard convolution layer (kernel size = 3×3), which expands the number of channels. Then it is followed by 6 DPD blocks and a global average pooling layer reducing the spatial resolution to 1 from 8×8. The final fully connected layer is fed into the softmax layer for classification. All convolution layers are followed by batch normalization [32] and RELU [39] nonlinear activation. The first DWC in DPD block can implement downsampling. The hyperparameter $\alpha$ can adjust the number of output channels of DPD block, which can make a trade-off between accuracy and computational complexity.

The base DPDNet architecture is very small with low latency. In many fields, high accuracy is required. So we introduce a very simple hyperparameter $\alpha$. We know wider models have better performance and $\alpha$ can widen DPDNet uniformly at each layer. Simultaneously, the number of parameters and computational cost are increased. In this paper, $\alpha$ is set as {1.25, 1.5, 1.75, 2.0, 2.5, 3.0, 4.0} and can be applied to any model structure to define a new wider model.

Furthermore, a larger DPDNet is constructed for more complex ImageNet [5] datasets, which is shown in Table 4. The main body architectures of MobileNetV2 and DPDNet are the same except the DPD block. Each line describes a sequence of 1 or more identical layers, repeated $n$ times. $s$ is the stride of convolution. $\alpha$ is set as 1 and m is set as 6.

Table 3. DPDNet Body Architecture for ImageNet.

| Input | Operator | *s* | *n* |
|---|---|---|---|
| 224×224×3 | Conv2d | 2 | 1 |
| 112×112×32 | DPD block | 1 | 1 |
| 112×112×16 | DPD block | 2 | 2 |
| 56×56×24 | DPD block | 2 | 3 |
| 28 ×28×32 | DPD block | 2 | 4 |
| 14 ×14×64 | DPD block | 1 | 3 |
| 14 ×14×96 | DPD block | 2 | 3 |
| 7×7×160 | DPD block | 1 | 1 |
| 7×7×320 | DPD block | 1 | 1 |
| 7×7×1280 | DPD block | - | 1 |
| 1×1×1280 | Avg pool 7 × 7 | 1 | - |
| 1×1×1000 | FC | - | - |

## 4.1. Datasets and Training Settings

**CIFAR.** The CIFAR [40] datasets, CIFAR-10 and CIFAR-100, are subsets of the 80 million tiny images [41]. The CIFAR-10 dataset consists of 10 classes, and each class contains 6000 color images of 32×32 with 5000 images for training and 1000 for testing. The CIFAR-100 dataset consists of 100 classes, and each class contains 600 color images of the same size with 500 images for training and 100 for testing. CIFAR-10 dataset has fewer classes and more images in each class. Therefore, CIFAR-10 dataset is simpler than CIFAR-100 dataset. The standard data augmentation [8, 9, 42-46] scheme we adopt is widely used for these datasets. We pad the images with 4 zero pixels on each side, and then randomly crop to produce 32×32 images, followed by horizontally mirroring half of the images. We normalize the images by using the channel means and standard deviations.

**CINIC-10.** CINIC-10 [47] consists of images from both CIFAR and ImageNet.

The CINIC-10 consists of 270000 32×32 images with 90000 images for training, 90000 for validation and 90000 for test. The train and validation subsets can be combined to make a larger training set. CINIC-10 dataset is larger and more challenging than CIFAR-10 but not as difficult as ImageNet. The standard data augmentation scheme we adopt is the same as CIFAR datasets.

**Training settings.** All models are trained in TensorFlow [48] using MomentumOptimizer algorithm to update network. The initial learning rate is 0.1 and multiplies with a factor 0.1 at 150 and 225 training epochs. The weight decay and the momentum are set as 0.0001 and 0.9 respectively. We train all networks for 300 epochs with a batch size of 128 on one GPU.

## 3.2. Results of PSDNet

We test the accuracies of PSDNet50 and compressed ResNet50 on CIFAR datasets. The comparison results of the number of parameters, computational costs and accuracies are given in Table 4. The comparison is fair because PSDNet50 and ResNet50 have similar structures, except that are just different in block structures. In a block of ResNet, the last layer is 1×1 convolution layer, which can combine information on different channels. In a PSD block, the last layer is DWC layer, which can extra spatial features on each channel.

The number of parameters of PSDNet50 is 0.4M less than the ResNet50, and PSDNet50 has less computational cost. Delightedly, the accuracy of PSDNet50 is 0.9% higher than ResNet50 on CIFAR-10 test dataset. Moreover, PSDNet50 outperforms the ResNet50 on CIFAR-100 dataset, and achieves about 0.5% accuracy improvement. Maybe the reason is that PSD block extracts more spatial features than ResNet block.

Table 4**.** Performance of PSDNet50 vs ResNet50 on CIFAR-10 (C-10) and CIFAR-100 (C-100). The #Params is the number of parameters of model for CIFAR-10.

| Network | #Params | FLOPs | C-10 | C-100 |
|---------|---------|-------|------|-------|
| ResNet50 | 2.0 M | 316 M | 92.95 | 72.61 |
| PSDNet50 | 1.6 M | 208 M | 93.87 | 73.14 |

## 3.3. Results of DPDNet

**The effect on m.** The hyperparameter $m$ makes a trade-off between accuracy and computational cost, and determines the multiple of the channel expansion in blocks. We analyze the effect of different m {1, 2, 3, 4, 5, 6} on accuracy. In order to verify the performance of DPDNet, we design a corresponding model MobileNetV2. The original MobileNetV2 is designed for ImageNet, which has 56 layers. In this paper, the MobileNetV2 is modified to 20 layers, and the modified MobileNetV2 has the same number of output channels of every block with DPDNet. The experimental results are shown in Table 5. The number of parameters in Table 5 is the model for CIFAR-10 and CINIC-10 datasets. The model for CIFAR-100 has larger number of parameters, because its FC layer has more output neurons. The convolutional layers for CIFAR-10, CINIC-10 and CIFAR-100 are the same, which occupies most of the number of parameters. As can be seen, DPDNet has fewer number of parameters than MobilenetV2 with the same $m$. The larger $m$ is, the larger the difference in the number of parameters is. The number of parameters of DPDNet is about $0.6 \times$ of MobileNetV2.

In CIFAR-10 dataset, the performance of DPDNet is significantly better than that of MobilenetV2 with the same value of hyperparameter $m$. As the value of $m$ decreases, this trend becomes more apparent. Only when $m = 4$, is the accuracy of MobilenetV2 slightly better than that of DPDNet.

In CIFAR-100 dataset, when $m = 1$, $2$, the performance of DPDNet is significantly better than that of MobilenetV2. Especially when $m = 1$, DPDNet is 5% higher than MobileNetV2 in accuracy. Another case is that the performance of MobileNetV2 and DPDNet is similar while $m > 2$. But DPDNet and MobilenetV2 share the same value of $m$, DPDNet has much fewer parameters than MobilenetV2, which suggests that the DPDNet is more efficient than MobileNetV2.

In CINIC-10 dataset, we only deploy the DPDNet. As we can see, as the value of m increases, the accuracy is higher. When $m > 4$, the accuracy is higher than 80%. DPDNet has good performance in CINIC-10 dataset.

Table 5. Performance on CIFAR-10 (C-10), CIFAR-100 (C-100) and CINIC-10 (CN-10) datasets of own implemented models at different $m$. Results of DPDNet that outperform MobileNetV2 at the same $m$ are **bold**.

| $m$ | Network | #Params | FLOPs | C-10 | C-100 | CN-10 |
|---|---|---|---|---|---|---|
| 1 | DPDNet | 0.04 M | 5.3M | **87.73** | **60.60** | 76.65 |
| | MobilenetV2 | 0.05 M | 8.3M | 86.19 | 55.39 | - |
| 2 | DPDNet | 0.06 M | 8.9M | **89.83** | **63.92** | 77.64 |
| | MobilenetV2 | 0.09 M | 15.8M | 89.10 | 62.99 | - |
| 3 | DPDNet | 0.09 M | 12.6M | **90.42** | 66.36 | 78.17 |
| | MobilenetV2 | 0.14 M | 23.3M | 89.92 | 67.07 | - |
| 4 | DPDNet | 0.12 M | 16.3M | 90.98 | 67.44 | 79.87 |
| | MobilenetV2 | 0.18 M | 30.8M | 91.09 | 68.01 | - |
| 5 | DPDNet | 0.15 M | 20.0M | **91.56** | 67.97 | 81.04 |
| | MobilenetV2 | 0.23 M | 38.2M | 91.45 | 69.35 | - |
| 6 | DPDNet | 0.17 M | 23.7M | **92.13** | 69.15 | 81.31 |
| | MobilenetV2 | 0.27 M | 45.7M | 91.97 | 69.66 | - |

**The effect on $\alpha$.** Another hyperparameter $\alpha$ also makes a trade-off between accuracy and computational cost, and can decrease or increase the number of channels of a block. In this paper, we test the performance of DPDNet when $m = 5$, $\alpha = \{1.25, 1.5, 1.75, 2.0, 2.5, 3.0, 4.0\}$. Similarly, we design the corresponding MobilenetV2 for each DPDNet. The experimental results are shown in Table 6. The number of parameters is the model for CIFAR-10 and CINIC-10 datasets.

As can be seen from Table 6, as $\alpha$ becomes larger, the number of parameters increases significantly. This is because the number of channels per layer becomes larger when $\alpha$ becomes larger. Nevertheless, when $m$ increases, only the number of first layer channels increases in a block.

In CIFAR-10 dataset, as the value of $\alpha$ increases, the performance of both DPDNet and MobilenetV2 becomes better. When $\alpha < 2$, the performance of DPDNet is better than that of MobilenetV2. However, when $\alpha > 2$, the results are opposite. When $\alpha$ is the same, DPDNet still has fewer number of parameters. We compared DPDNet and MobilenetV2 in terms of number of parameters and accuracy. It can be seen that when the number of parameters are similar, the performance of DPDNet is better than that of MobilenetV2.

**Table 6.** Performance on CIFAR-10（C-10）, CIFAR-100 (C-100) and CINIC-10 (CN-10) datasets of own implemented models at different $\alpha$. Results of DPDNet that outperform MobileNetV2 at the same $\alpha$ are **bold**.

| $\alpha$ | Network | #Params | FLOPs | C-10 | C-100 | CN-10 |
|------|------------|---------|-------|-------|-------|-------|
| 1.25 | DPDNet | 0.22 M | 28.3M | **92.03** | 69.26 | 81.45 |
|      | MobilenetV2 | 0.34 M | 54.1M | 91.93 | 70.73 | - |
| 1.5  | DPDNet | 0.31 M | 38.1M | **92.19** | 70.66 | 81.25 |
|      | MobilenetV2 | 0.49 M | 73.2M | 91.61 | 71.59 | - |
| 1.75 | DPDNet | 0.42 M | 49.5M | **92.14** | **71.99** | 81.63 |
|      | MobilenetV2 | 0.65 M | 95.6M | 91.86 | 71.77 | - |
| 2.0  | DPDNet | 0.54 M | 62.5M | **92.23** | **72.48** | 81.66 |
|      | MobilenetV2 | 0.85 M | 121M | 92.07 | 72.21 | - |
| 2.5  | DPDNet | 0.83 M | 93.1M | 92.41 | **73.72** | 82.72 |
|      | MobilenetV2 | 1.30 M | 182M | 92.47 | 73.32 | - |
| 3.0  | DPDNet | 1.18 M | 130M | 92.66 | **73.78** | 83.24 |
|      | MobilenetV2 | 1.86 M | 257M | 92.72 | 73.63 | - |
| 4.0  | DPDNet | 2.07 M | 222M | 92.86 | **74.75** | 84.05 |
|      | MobilenetV2 | 3.28 M | 444M | 93.27 | 74.17 | - |

In CIFAR-100 dataset, when $\alpha < 1.75$, the accuracy of MobilenetV2 is better than DPDNet. But when $\alpha > 1.75$, situation is the opposite. By comparing the relationship between parameters and accuracy, we can see that DPDNet has higher accuracy than MobilenetV2 when the number of parameters is almost the same.

In CINIC-10 dataset, the accuracy of DPDNet becomes higher as $\alpha$ increases. Darlow et al. reported that the accuracy of MobileNetV2 with 3.2M parameters on CINIC-10 is 82% [47]. However, the accuracy of DPDNet ($\alpha = 2.5$) with 0.83 M parameters is 82.72%, which indicates DPDNet is more efficient than MobileNetV2.

In ImageNet, our proposed DPDNet has  M parameters and  M FLOPs. Table x shows the accuracies of DPDNet and some other efficient lightweight CNNs on ImageNet. Firstly, we can see that parameters and FLOPs of DPDNet are M and M fewer than that of MobileNetV2, respectively.

Furthermore, the DPDNet ($\alpha = 1.25$, $m = 5$) has large number of parameters than DPDNet ($\alpha = 1.0$, $m = 6$). Whereas, the accuracy of DPDNet ($\alpha = 1.25$, $m = 5$) is lower than DPDNet ($\alpha = 1.0$, $m = 6$) on CIFAR-10 and CIFAR-100 datasets, which suggests that m may be more effective in improving the performance.

**Table 8.** Performance of DPDNet and some other lightweight CNNs on ImageNet.

| Network | #Params | FLOPs | TOP-1 |
|---|---|---|---|
| DPDNet | | | |
| MobileNetV2 | 3.4 M | 300 M | 72.0 |
| MobileNetV1 | 4.2 M | 575 M | 70.6 |
| ShuffleNet (1.5×) | 3.4 M | 292 M | 71.5 |
| ShuffleNetV2 (1.5×) | - | 299 M | 72.6 |
| IGCV3-D | 3.5 M | 318 M | 72.2 |
| Xception (1.5×) | - | 305 M | 70.6 |

Moreover, there are one DWC layer and two PWC layers in a block of MobileNetV2. There are two DWC layers and a PWC layer in a block of DPDNet. DPDNet block extracts more spatial features and MobileNetV2 block assembles more information on different channels. DPDNet has fewer number of parameters and achieves similar or better performance compared with MobileNetV2. The results suggest that extracting spatial features may be more important in CNNs.

4. Conclusion

This paper proposes using DWC to increase the number of channels instead of common PWC. By deploying DWC layer to expand the number of channels, we introduce an efficient CNN architecture, which is called PSDNet with fewer number of parameters, less computational cost and higher accuracy than the corresponding ResNet. In order to design lightweight CNN, the PDP inversed bottleneck block and DPDNet are proposed, which has more efficient DWC layers. And DPDNet is more efficient than MobileNetV2. The computational cost and performance of DPDNet can be controlled by adjusting the width multiplier $\alpha$ and the channel-multiplier $m$. In addition, we find that the CNNs extracting more spatial features have higher accuracy than ones combining information among channels, which indicates spatial features are more important in CNNs.

In the future, it is necessary to evaluate deeper DPDNet with more experiments on the ImageNet dataset.

**Acknowledgment**

This research work was partly supported by the Natural Science Foundation of China and Jiangsu (Project No. 61750110529, 61850410535, 61671148, BK20161147), the Research and Innovation Program for Graduate Students in Universities of Jiangsu Province (Grant No. SJCX17_0048, SJCX18_0058).